\documentclass[11pt]{article} % For LaTeX2e
\usepackage{rldmsubmit,palatino}
\usepackage{graphicx}
\usepackage{caption}
\usepackage{subcaption}
\usepackage{natbib}
\usepackage[capitalize,noabbrev]{cleveref}
\usepackage{url}
\usepackage{wrapfig}

\def\hc{\texttt{HalfCheetah}}
\def\ww{\texttt{Walker2d}}
\def\ant{\texttt{Ant}}
\def\hum{\texttt{Humanoid}}

\makeatletter
\newcommand{\printfnsymbol}[1]{%
  \textsuperscript{\@fnsymbol{#1}}%
}
\makeatother

\title{PixelBrax: Learning Continuous Control from Pixels End-to-End on the GPU} % max 100 chars

\author{
Trevor McInroe\thanks{Equal contribution. Correspondence to \texttt{\{t.mcinroe,s.garcin\}@ed.ac.uk}} \\
The University of Edinburgh \\
Edinburgh, UK \\
\texttt{t.mcinroe@ed.ac.uk}
\And
Samuel Garcin\printfnsymbol{1} \\ 
The University of Edinburgh \\ 
Edinburgh, UK \\
\texttt{s.garcin@ed.ac.uk}
}

\begin{document}

\maketitle

% base:
% We present PixelBrax, a set of reinforcement learning (RL) environments with pixel-based observation spaces that runs end-to-end on the GPU. By combining the Brax~\citep{braxarxiv} physics engine with a pure JAX renderer and JAX-accelerated neural network training, the RL training loop \textit{including the environment step} can be compiled. We provide steps-per-second benchmark figures for PixelBrax that demonstrate how well it scales with even tens of thousands of parallel environments. Also, we provide example returns for several RL algorithms that take advantage of environment parallelism. PixelBrax currently supports adding color and video distractors. In the future, we plan to increase the number of supported environments, provide a comprehensive benchmark of RL algorithms, and add support for mesh rendering.

\begin{abstract} % max 2000 chars including spaces
We present PixelBrax, a set of continuous control tasks with pixel observations.  We combine the Brax physics engine with a pure JAX renderer, allowing reinforcement learning (RL) experiments to run end-to-end on the GPU. PixelBrax can render observations over thousands of parallel environments and can run two orders of magnitude faster than existing benchmarks that rely on CPU-based rendering. Additionally, PixelBrax supports fully reproducible experiments through its explicit handling of any stochasticity within the environments and supports color and video distractors for benchmarking generalization. We open-source PixelBrax alongside JAX implementations of several RL algorithms at \url{github.com/trevormcinroe/pixelbrax}. %In the future, we plan to increase the number of supported environments, provide a comprehensive benchmark of RL algorithms, and add support for mesh rendering.
\end{abstract}


% \acknowledgements{We are deeply indebted to Google DeepMind and
% the Weinber?g Institute for Cognitive Science for
% their generous support of RLDM2017.}  

\startmain % to start the main 1-4 pages of the submission.

\section{Introduction}
Combining JAX-accelerated neural network training and JAX-accelerated reinforcement learning (RL) environments such as Brax~\citep{braxarxiv} massively shortens the wall-clock time required to train RL agents. This speedup occurs because the training loop \textit{including the environment step} runs end-to-end on the GPU, which allows massive scaling of the number of environments that can run in parallel. However, Brax does not currently support on-GPU rendering.

Brax's built-in renderer relies on utilities provided by MuJoCo~\citep{mujoco}, which uses CPU-based rendering. Thus, using the current Brax rendering utilities would drastically reduce throughput by forcing data to transit between the CPU and GPU every timestep. Given the popularity of RL research with pixel observations~\citep{DQN,dreamerv1,mhairiCMID,Garcin2024DREDZT}, the community would benefit from high-throughput RL environments with pixel states.

We present PixelBrax, a set of RL environments with pixel-based observation spaces that run end-to-end on the GPU. Using open-source utilities~\citep{jaxrenderer}, PixelBrax delegates environment rendering to the accelerator via JAX, which ultimately allows end-to-end GPU online RL training from pixels over thousands of environments in parallel. The user has the option of enabling color and video distractors with little to no additional computational overhead. Experiments conducted in PixelBrax are fully reproducible given a random seed, as PixelBrax handles any stochasticity within the environment and distractors via JAX's explicit pseudorandom number generation.

\section{PixelBrax}
\begin{figure}[!t]
     \centering
     \begin{subfigure}[b]{0.3\textwidth}
         \centering
         \includegraphics[width=\textwidth,angle=90,origin=c]{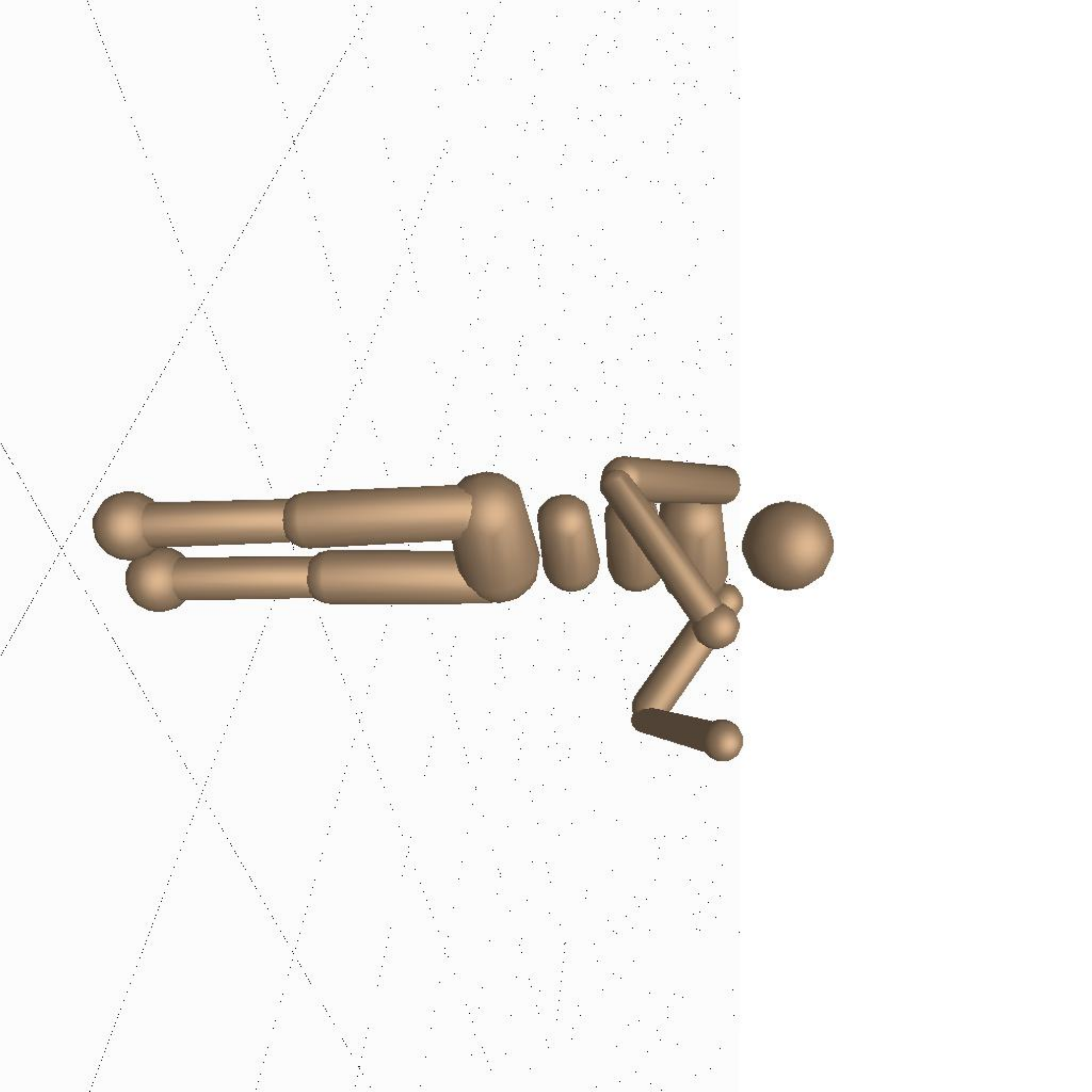}
         % \caption{$y=x$}
         % \label{fig:y equals x}
     \end{subfigure}
     \hfill
     \begin{subfigure}[b]{0.3\textwidth}
         \centering
         \includegraphics[width=\textwidth,angle=90,origin=c]{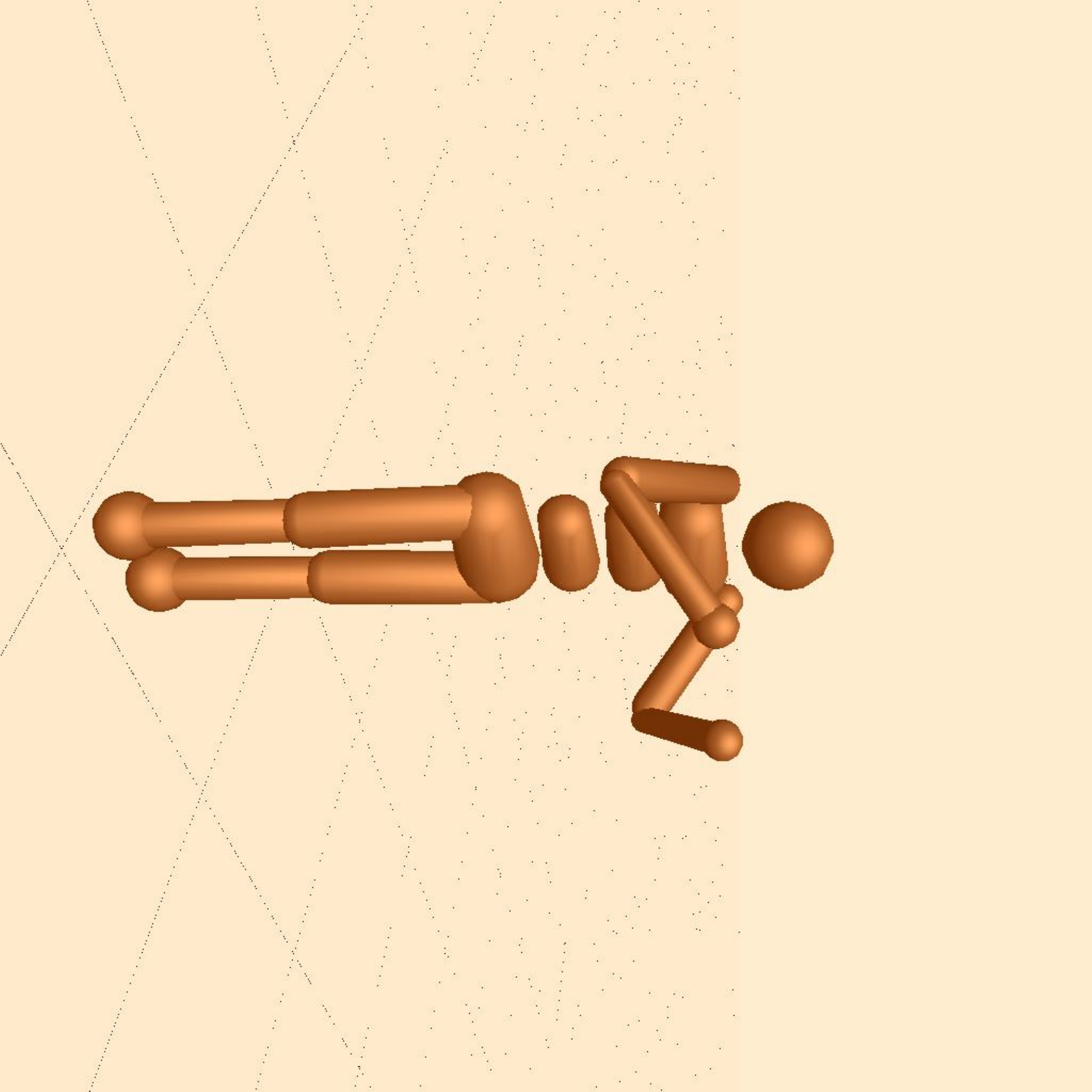}
         % \caption{$y=3\sin x$}
         % \label{fig:three sin x}
     \end{subfigure}
     \hfill
     \begin{subfigure}[b]{0.3\textwidth}
         \centering
         \includegraphics[width=\textwidth,angle=90,origin=c]{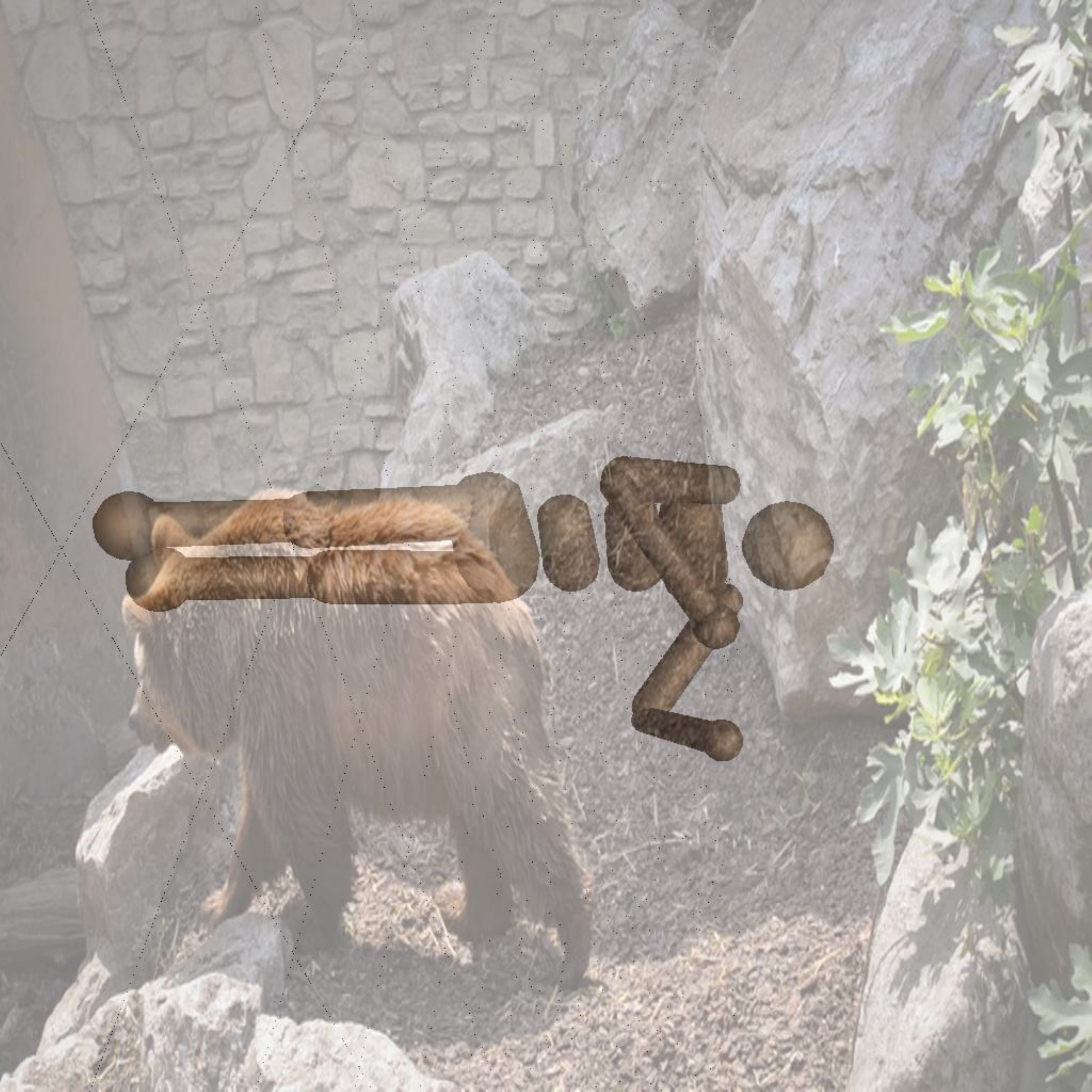}
         % \caption{$y=5/x$}
         % \label{fig:five over x}
     \end{subfigure}
        \caption{An initial state in the \hum $\,$ environment with no distractors, color distractors, and video distractors (left-to-right). When using color distractors, the color adjustment to each frame is different at each timestep. Likewise, when using video distractors, each subsequent timesteps' pixels are overlayed with the subsequent frame in the video.}
        \label{fig:humanoid}
\end{figure}

% \begin{wrapfigure}[]{R}{0.5\textwidth}
% \vspace{-1cm}
%     % \centering
%     % \begin{subfigure}[]{1\linewidth}
%         \includegraphics[width=0.98\linewidth]{figs/sps_pb.pdf}
%     % \end{subfigure}
%     \caption{Steps per second for all four currently-supported environments over $\{ 1, 10, 100, 1000 \}$ parallel environments}\label{fig:sps}
%     % \vspace{500cm}
% \end{wrapfigure}

PixelBrax currently supports four Brax environments: \hc, \ant, \ww, and \hum. In each environment, the user may enable ``color" or ``video" distractors (see ~\Cref{fig:humanoid}). Adding color distractors causes an entire image's pixel's color-bias to change randomly at each environment step. When using video distractors, a random video from the Davis-2017~\citep{davis} dataset is played on top of the environment's pixels, where subsequent environment steps' pixels are overlayed with subsequent video frames. PixelBrax implements distractors directly within the JAX renderer, and, for video distractors, will load the full video into GPU memory. We report no noticeable reduction in rendering and simulation speed when enabling distractors. In contrast, prior implementations of video distractors~\citep{stone2021distracting} load individual frames from disk into the MuJoCo simulator, causing a significant slowdown in the number of simulated steps per second. Finally, PixelBrax relies on JAX's explicit handling of random number generation to handle any stochasticity in the environment dynamics or distractors. This makes reproducing individual trajectory rollouts (or an entire RL experiment) straightforward.

% This contrasts with other pixel-based environments with distractors that rely on NumPy, like Distracting Control Suite~\citep{DistractingDM}, where controlling randomness in sampling is not straightforward. 

\begin{figure}[h]
    \centering
    \includegraphics[width=0.45\linewidth]{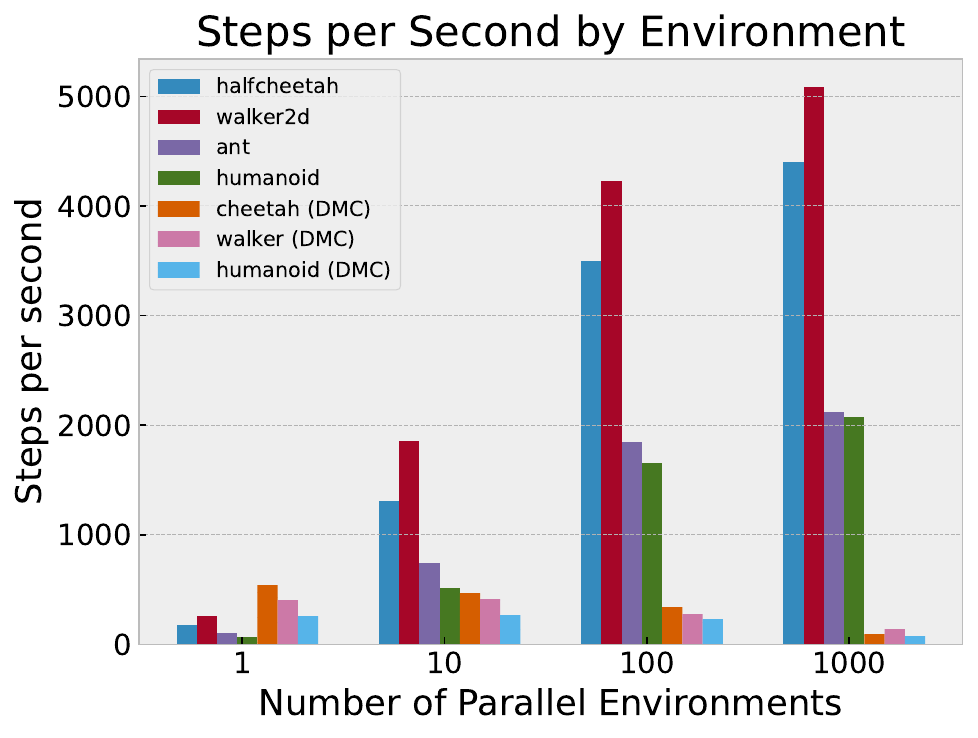}
    \caption{Steps per second for all four currently-supported PixelBrax environments (non-hashed) and three DMC from pixels environments (hashed) over $\{ 1, 10, 100, 1000 \}$ parallel environments.}
    \label{fig:sps}
\end{figure}
\Cref{fig:sps} provides throughput measured in steps per second for each PixelBrax environment and three environments from the DeepMind Control Suite (DMC) over a number of parallel environments in $\{ 1, 10, 100, 1000 \}$. The DMC environments are rendered with dmc2gym~\citep{dmc2gym}, a popular wrapper for DMC from pixels, and vectorized with stable-basedlines3~\citep{sb3}. Each environment step in the loop includes a forward pass through a convolutional layer from Flax~\citep{flax} to simulate selecting actions with the agent's policy. We ran this benchmark using one 40GB A100 and AMD EPYC 7713 64-Core Processor. We note that the number of steps-per-second in PixelBrax scales well with the number of parallel environments. In contrast, the DMC environments' throughput worsens as the number of parallel environments increases.

\begin{figure}[!h]
     \centering
     \begin{subfigure}[b]{0.35\textwidth}
         \centering
         \includegraphics[width=\linewidth]{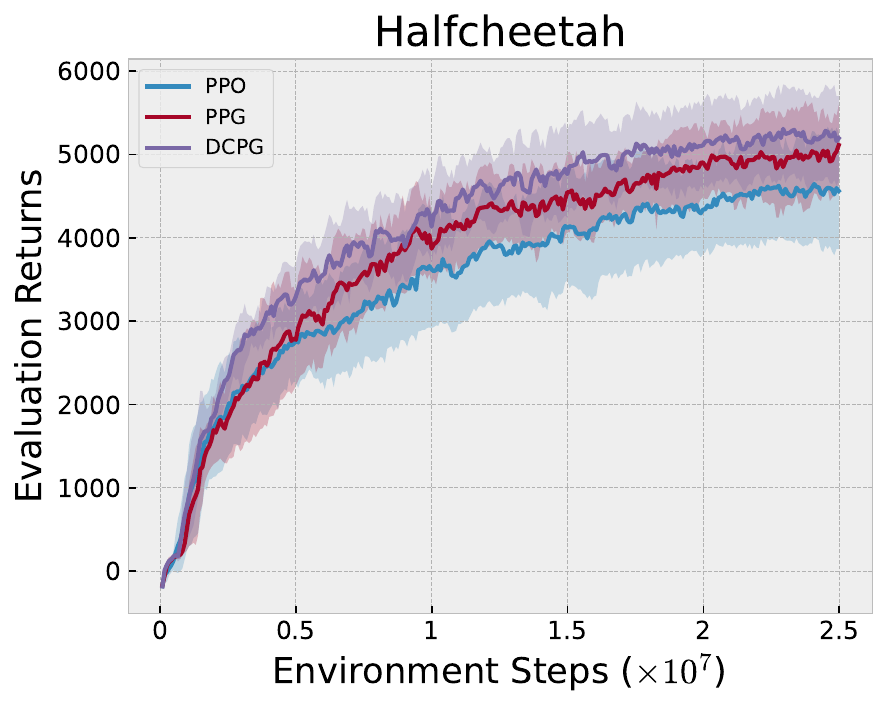}
         % \caption{$y=x$}
         % \label{fig:y equals x}
     \end{subfigure}
     % \hfill
     \begin{subfigure}[b]{0.35\textwidth}
         \centering
         \includegraphics[width=\textwidth]{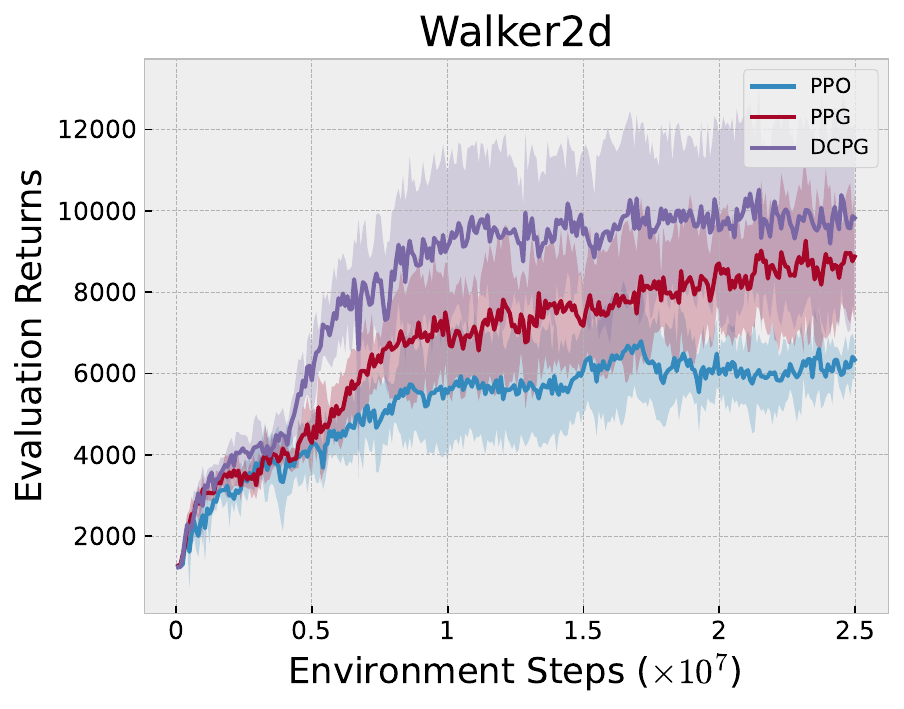}
         % \caption{$y=x$}
         % \label{fig:y equals x}
     \end{subfigure}

     \begin{subfigure}[b]{0.35\textwidth}
         \centering
         \includegraphics[width=\textwidth]{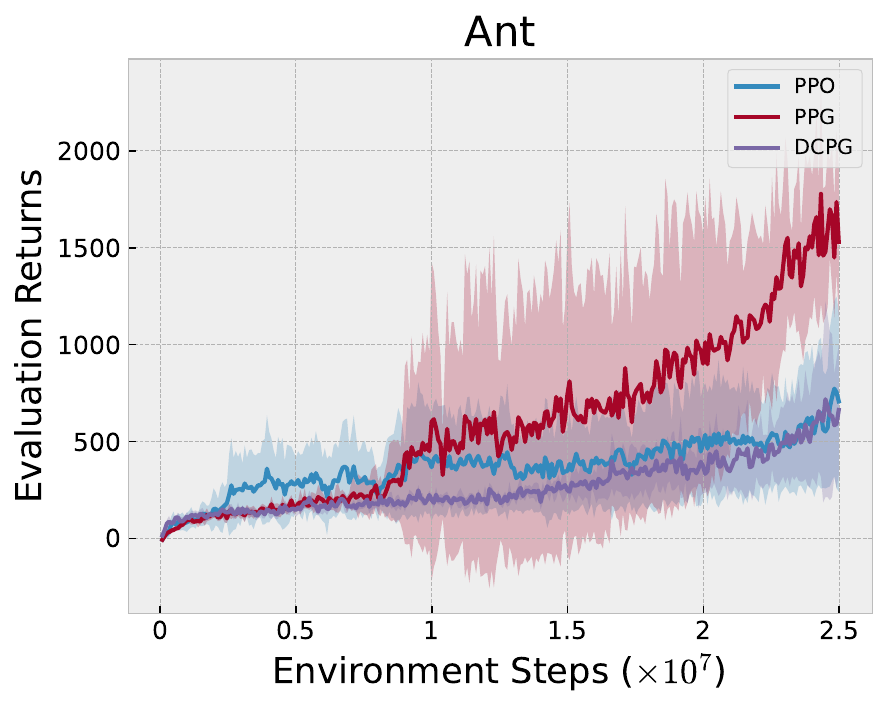}
         % \caption{$y=x$}
         % \label{fig:y equals x}
     \end{subfigure}
     % \hfill
     \begin{subfigure}[b]{0.35\textwidth}
         \centering
         \includegraphics[width=\textwidth]{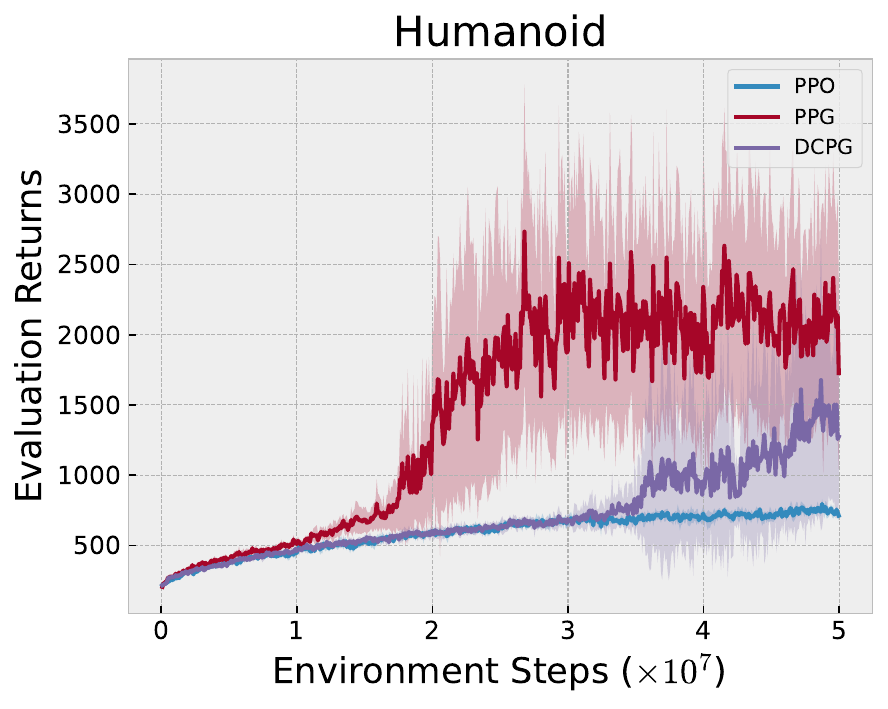}
         % \caption{$y=x$}
         % \label{fig:y equals x}
     \end{subfigure}
        \caption{Example returns for PPO, PPG, and DCPG in the \hc $\,$ (top-left), \ww $\,$ (top-right), \ant $\,$ (bottom-left), and \hum $\,$ (bottom-right), all with video distractors. Bold line represent mean, shaded area represents $\pm$ one standard deviation.}
        \label{fig:video-bench}
\end{figure}

In~\Cref{fig:video-bench}, we provide example returns over five seeds with video distractors for PPO~\citep{PPO}, PPG~\citep{PPG}, and DCPG~\citep{DCPG} in \hc, $\,$ \ww, $\,$ and \ant $\,$ over 25 million timesteps, and in \hum $\,$ over 50 million timesteps.

% \section{Future Plans}
% We plan to add more environments, a comprehensive benchmark over many RL algorithms, and support for mesh rendering in the future.

\bibliographystyle{unsrtnat}
\bibliography{bib.bib}

\end{document}